\documentclass{article}

\usepackage{PRIMEarxiv}

\usepackage[utf8]{inputenc} 
\usepackage[T1]{fontenc}    
\usepackage{hyperref}       
\usepackage{url}            
\usepackage{booktabs}       
\usepackage{amsfonts}       
\usepackage{nicefrac}       
\usepackage{microtype}      
\usepackage{lipsum}
\usepackage{microtype}
\usepackage{graphicx}
\usepackage{placeins}
\usepackage{subcaption}
\usepackage{mathtools}

\usepackage{hyperref}
\usepackage{fancyhdr}       
\usepackage{graphicx}       
\graphicspath{{media/}}     

\title{Conformalised imprecise inference for robust extrapolation under limited data
}

\author{
  Yu Chen\\
  Institute for Risk and Uncertainty \\
  University of Liverpool \\
  Liverpool\\
  \texttt{yuchen2@liverpool.ac.uk} \\
   \And
  Scott Ferson \\
  Institute for Risk and Uncertainty \\
  University of Liverpool \\
  Liverpool\\
  \texttt{sandp8@gmail.com} \\
}

\begin{document}
\maketitle

\begin{abstract}
Recent advances in uncertainty quantification increasingly emphasise the distinction between aleatory and epistemic uncertainty in machine learning, motivating the need for more unified frameworks. However, despite much progress in producing reliable predictions, existing methods often lack rigorous guarantees when generalising beyond the training domain. We propose a conformalised imprecise inference framework for robust extrapolation, which is model-agnostic and augments predictive models with imprecision and distance awareness. The proposed approach yields  imprecise predictions (probability boxes) that remain valid under distributional shift, maintaining coverage while adaptively expanding uncertainty in extrapolation regimes. Experiments on synthetic and benchmark datasets demonstrate improved robustness and reliable coverage compared to standard probabilistic approaches, particularly under limited data.
\end{abstract}


\section{Introduction}

Generalising beyond the data seen from learning is an essential task of machine learning systems \cite{hullermeier2021aleatoric}. 
Much progress in uncertainty quantification (UQ) for artificial intelligence (AI) has focused on addressing the overconfidence of trained models in high-stakes settings \cite{gal2016dropout, liu2020simple, romano2019conformalized,hullermeier2021aleatoric}.
Importantly, recent work has increasingly distinguished between aleatory and epistemic sources of predictive uncertainty,  with particular emphasis on the latter, focusing on challenges in limited data, measurement noise, model assumptions, distributional shift, and extrapolation \cite{hullermeier2014learning, destercke2022uncertain, tretiak2023neural}.

Modern theories of uncertainty, e.g. the imprecise probability framework such as credal sets, evidence theory (Dempster Shafer structures), hierarchical higher-order distribution, uncertain numbers, fuzzy sets etc \cite{dempster1968generalization, destercke2008unifying, chen2025}, can represent much weaker statements of knowledge and more diverse types of uncertainty than conventional probability theory to reason under polymorphic uncertainty. 
Following the rapid development of approximate Bayesian inference in deep learning, these modern theories are applied to comprehensively analyse the uncertainty landscape of the machine learning systems \cite{caprio2023credal, hofman2024quantifying, amini2020deep, wang2024credal}.  

The pursuit of uncertainty research align with the goal of enhancing the robustness of the AI systems, wherein statistical guarantee for generalisation is desired, even with distribution shift.
Epistemic learning theory \cite{manchingal2025epistemic} emphasizes that the system must be designed to be prepared for data it has not yet encountered. Despite much progress towards the guarantee on the generalisation performance, with notable examples including PAC-Bayesian theory \cite{germain2016pac} and classic conformal prediction \cite{romano2019conformalized}, it is challenging to produce rigorous bounds that are guaranteed to enclose the ground truth under distributional shift, particularly in extrapolation settings.

In this work, we investigate epistemic uncertainty in extrapolation settings, while systematically analysing the impact of limited data through controlled experiments. We propose a conformalised imprecise inference (CII) framework that integrates distance awareness with imprecise probability representations. In particular, we leverage a scalar distance score to characterise deviation from the training data manifold and use it to drive a distance-dependent discrepancy model. This discrepancy is embedded within a novel \textit{ascloseas} bounds construction, yielding conformalised p-box predictions that maintain coverage while adaptively expanding uncertainty under distributional shift.

\section{Conformalised framework for robust extrapolation}

\subsection{Background on predictive capability with uncertainty modelling}
\label{sec:uq_context}

Consider a general learning problem in a regression setting. Given  observation data $\mathcal{D} = \{(\mathbf{x}_{i}, y_{i})_{i=1}^{N}\} \in (\mathcal{X} \times {\mathcal{Y}})$ which are assumed to independent and identically distributed (i.i.d) with a distribution $\mathcal{P}_{D}$, the expected loss of a hypothesis $h \in \mathcal{H}: \mathcal{X} \rightarrow \mathcal{Y}$ is given as $\mathcal{L} = \mathbb{E}_{(\mathbf{x}, y) \sim \mathcal{P}_{D}} \big [\ell ({\mathbf{x}}, y) \big ]$.

To formulate heteroscedastic aleatory uncertainty, neural networks can be viewed as probabilistic models with a data-dependent noise structure
$p(y_i \mid f^{\boldsymbol{\omega}}(\mathbf{x}_i)) =
\mathcal{N}\left(\mu(\mathbf{x}_i), \sigma^2(\mathbf{x}_i)\right)$.
The empirical loss, given by the negative log-likelihood, is written as:

\begin{equation}
\mathcal{L}^{AU} = \frac{1}{N} \sum_{i=1}^{N} \frac{1}{2 \sigma(\mathbf{x}_{i})^2} (y - \mu(\mathbf{x}_{i}))^2 + \frac{1}{2} \log \sigma(\mathbf{x}_{i})^2
\end{equation}

To model the epistemic uncertainty on model parameters, Bayesian deep learning evaluates the posterior weights $p(\boldsymbol{\omega}|\mathcal{D})$ given the observation data, see xx for variants of approximate Bayesian inference to handle the otherwise intractable inference problem. Notably,  SVI turns the inference problem into optimisation of a stochastic objective named ELBO by introducing variational distribution $q_{\phi}(\boldsymbol{\omega})$ parameterised by $\phi$:

\begin{equation}
	\mathcal{L}^{EU} = \text{KL} [q_{\phi}(\boldsymbol{\omega}) \  \| \ p(\boldsymbol{\omega})] - \mathbb{E}_{q_{\phi}(\boldsymbol{\omega})} \log p(\mathcal{D}|\boldsymbol{\omega}) \label{eq:nagative ELBO}
\end{equation}


In compounding these two sources, researches differ in a mixture route to get compound distribution \cite{kendall2017uncertainties, lakshminarayanan2017simple} or a credal route to obtain imprecise structures \cite{caprio2023credal, hofman2024quantifying}. In addition, conformal prediction (CP)\cite{hoff2023bayes} presents as a new paradigm for quantifying predictive uncertainty by enriching a prediction from a  hypothesis $h$ into prediction sets.

\subsection{Conformalised inference with distance awareness and distributional imprecision}

Compared to the classic CP where conformity guarantee is constructed on the assumption of \textit{exchangeability} which may be violated under distributional shift, we proposed an extrapolation-focused approach where in-distribution (ID) and out-of-distribution (OOD) data are explicitly differentiated, in principle allowing the robustness against extrapolation problems on the basis of distributional awareness. Meanwhile, the proposed conformalised framework is also agnostic to any pre-trained model which may  possess some capacity of uncertainty quantification.

With respect to the trans-probabilistic predictive distributions yielded by the base hyposis, in which the predictive uncertainty could be manifested in different uncertainty models: probability distributions, a set of distributions (e.g. p-boxes), or intervals; whereas the observation may be precise or subject to measurement uncertainty. We employ the \textit{stochastic area metric} as the conformal score, $d(x)$, suggesting the discrepancy between two random data generating processes, the predictive model $p(y|\mathbf{x})$ and the real-world observation $\mathcal{Y}_{obs}$, under polymorphic uncertainty. That is, in recognising the measurement uncertainty, the observation data could be elicited as intervals (say plus-minus form) or distributions (say PERT distribution).
Generally, these are effectively uncertain numbers \cite{chen2025} at varying level of imprecision, and the stochastic area metric \cite{ferson2008model} in a generalised formulation is given by:

\begin{equation}
	d(x) \coloneqq \mathbb{E}_x \Delta([\underline{F}(x), \overline{F}(x)], [\underline{S}(x), \overline{S}(x)])
\end{equation}

where $[\underline{F}(x), \overline{F}(x)]$ represent the bounds of the probability box of the prediction, while $[\underline{S}(x), \overline{S}(x)]$ represent the empirical probability box from the data.


Distance awareness has been identified as a key heuristic to reflect distribution shift \cite{liu2020simple}. To characterise extrapolation in the covariate space, we project inputs onto a distance space $\mathcal{R} \in \mathbb{R}$ embedding the deviation of a testing data point from the training data manifold. Specifically, We define the distance score as: $	r(\mathbf{x}) = \sqrt{(\mathbf{x}-\boldsymbol{\mu})^{T} \Sigma^{-1} (\mathbf{x}-\boldsymbol{\mu})}$, which corresponds to the Mahalanobis distance. When $\Sigma=I$, this reduces to the Euclidean distance. The score $r(x)$ induces an ordering over inputs that reflect their degree to distributional shift related to the training data.


To learn how predictive uncertainty evolves with distance, we construct a calibration subset $\mathcal{D}_c \subseteq \mathcal{D}$ in a hybrid manner that balances coverage of the input space using distance with an emphasis on boundary regions using the scenario optimisation \cite{de2021constructing}. 
Conformal scores are computed in $\mathcal{D}_c$ whereby an interval predictor model (IPM) $\mathcal{M}: r \rightarrow \mathcal{I}_d$, is constructed to rigorously characterise the range $\mathcal{I}_d = [\underline{d}, \overline{d}]$ for unseen inputs for generalisation. It benefits from the statistical guarantee of high probability of covering the unseen response \cite{campi2009interval}.

\begin{equation}
	\mathcal{I}_d(r, \mathcal{P}) = [\underline{d}, \overline{d}] = \{ f_l(r) \leq d(r) \leq f_u(r)\}
\end{equation}

where $\mathcal{M} = \{d(r) = M(r, \boldsymbol{q}), \boldsymbol{q} \in \mathcal{P} \}$ encodes a set of models parameterised by a vector $\boldsymbol{q}$ ranging in the set $\mathcal{P}$. Practically it is prescribed by its upper and lower boundaries $f_u(r) = \sum_{j=0}^{n_{u}} u_{j} \psi_j (r)$ and $f_l(r) = \sum_{j=0}^{n_{l}} l_{j} \psi_j (r)$, polynomial functions in which $\psi_{j}(r)$ are basis function and $n_{u}$ is the number of coefficients, that fully enclose the response with the minimal spread. This can be solved by a semidefinite program (SDP) with sum-of-squares (SOS) constraints \cite{lacerda2017interval} which even allows the distance to be imprecise.

We further adopt a distance-aware adjustment rule to reflect a transition from  accuracy-oriented in-distribution to conservative uncertainty inflation under extrapolation.
Let $r_0 = \sup_{x \in \mathcal{D}_{\mathrm{train}}} r(x)$ denote the boundary 
of the training support. We define:

\begin{equation}
\tilde{d}(x) =
\begin{cases}
\dfrac{d_l(r(x)) + d_u(r(x))}{2}, & r(x) \leq r_0, \\
d_u(r(x)), & r(x) > r_0,
\end{cases}
\end{equation}


Our approach allows the conformal adjustment to depend on the distance score $r(x)$, enabling a distance-aware calibration. 
As a result, the method retains reliable coverage in the core region while adaptively inflating uncertainty in extrapolation regimes. This provides a principled mechanism to bridge conformal prediction with distribution shift, without requiring explicit modelling of the test distribution.

\subsection{Ascloseas bounds with imprecision}


We recover the set of all the probability distributions within the area metric $d$ to the prediction of the base hypothesis. It is seeking the credal set of all possible predictions, the ball, under the constraint of given discrepancy measure:

\begin{equation}
	Q \in \mathcal{B}_{p}(P, Q) = \{ Q: W_p(P, Q) \leq d\}
\end{equation}

We re-interpret the area metric as the Wasserstein distance and thus exploiting an optimal mass transport plan given the discrepancy budget $d$. The envelope can be obtained:

\begin{equation}
	\overline{G}(x) = \sup_{Q: W_{1}(P, Q) \leq d, Q \in \mathcal{C}} Q((- \infty, x])
\label{eq:upper_asclose_bound}
\end{equation}

\begin{equation}
	\underline{G}(x) = \inf_{Q: W_{1}(P, Q) \leq d, Q \in \mathcal{C}} Q((- \infty, x])
\label{eq:lower_asclose_bound}
\end{equation}

where $\mathcal{C}$ encodes extra information on the admissible distributions in the ball, such as the support, moment, or the shape constraints. This aids in reducing the epistemic uncertainty of $Q$, serving as additional constraints for the optimisation programs of the Eq.~(\ref{eq:upper_asclose_bound}) and Eq.~(\ref{eq:lower_asclose_bound}).

\begin{figure*}[t]
     \centering
       \begin{subfigure}[b]{0.22\textwidth}
         \centering
         \includegraphics[width=\textwidth]{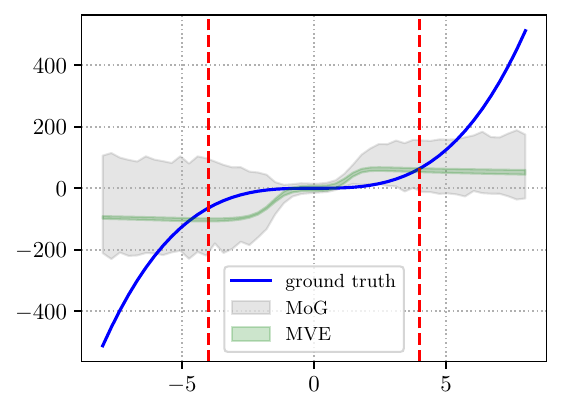}
         \caption{MVE and MoG}
         \label{fig:MVE_MoG_cubic}
     \end{subfigure}
     \begin{subfigure}[b]{0.22\textwidth}
         \centering
         \includegraphics[width=\textwidth]{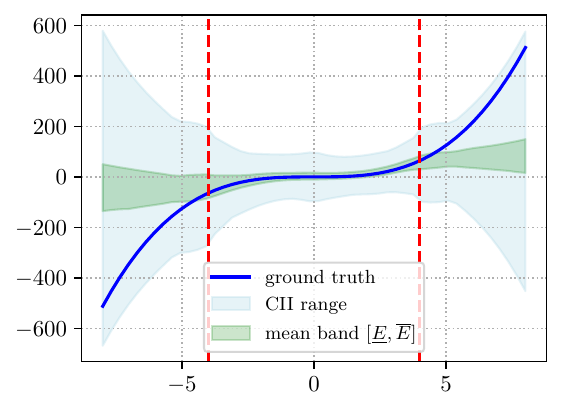}
         \caption{CII bands}
         \label{fig:CII_bands_cubic}
     \end{subfigure}
    \hspace{0pt}
     \begin{subfigure}[b]{0.22\textwidth}
         \centering
         \includegraphics[width=\textwidth]{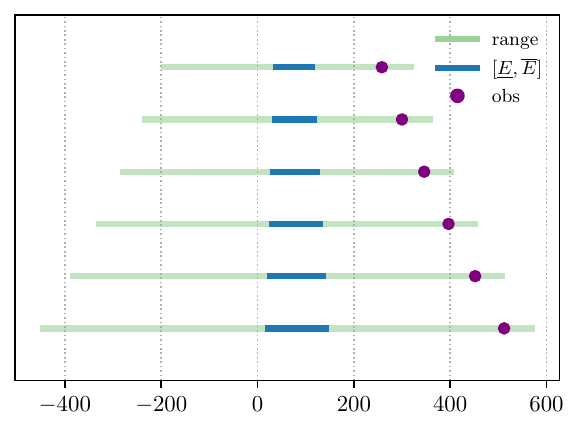}
         \caption{Range and mean interval}
         \label{fig:range_mean_band}
     \end{subfigure}
    \hspace{0pt}
     \begin{subfigure}[b]{0.22\textwidth}
         \centering
         \includegraphics[width=\textwidth]{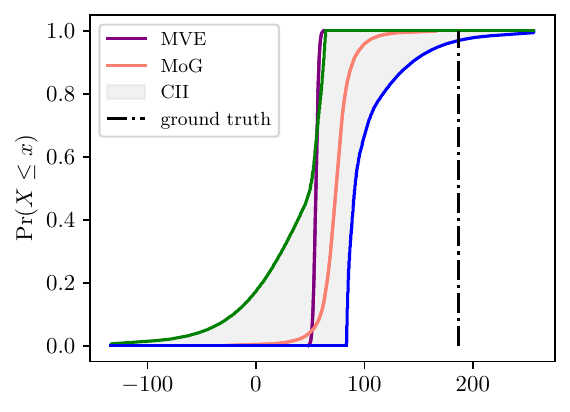}
         \caption{Predictive uncertainty}
         \label{fig:pbox_pred_ex_cubic}
     \end{subfigure}
\caption{Demonstration of the predictive capability on extrapolation with a toy example of the cubic function.}
\label{fig:cubic_toy_example}
\end{figure*}

\subsection{Conformalised imprecise inference}

Ideally, a reliable prediction should capture all the possible sources of uncertainty into the final prediction. 
Probability boxes, turns out to be a useful structure to confound various sources of uncertainty through the mechanism of \textit{probability bounding}, where its precision (i.e. the breadth $\gamma$ between the upper and lower edge) suggests the level of epistemic uncertainty. More importantly, the bounds can be accordingly inflated to account for additional aspects of uncertainty or approximation, serving as a principled way to aggregate uncertainties in a unified structure.

Based on the \textit{ascloseas} bounds, our method enriches a base prediction with imprecision and robustness to generalisation, as indicated below: 

\begin{equation}
	\mathcal{C}(x_{*}) = \{ y: d(x_{*}, y) \leq \tilde{d}(x_{*}) \} \sim [\underline{G}(x), \overline{G}(x)]
	\label{eq:CII_framework}
\end{equation}

As mentioned before, this framework accepts any form of the base prediction as an uncertain number which may be a distribution, a p-box or an interval, possibly yielded from the models in Section~\ref{sec:uq_context}.
To assess the predictive performance of the conformalised imprecise inference (CII), we consider three complementary metrics capturing accuracy, validity, and imprecision.

\textbf{Distance-stratified coverage.} 
We define the empirical coverage rate as:

\begin{equation}
	\xi = \frac{1}{N} \sum_{i=1}^N \mathbb{I}\{y_i \in C(x_i)\}
\end{equation}

which measures the proportion of observations enclosed by the predictive set. A reliable method should achieve high coverage, particularly under distributional shift.
To assess conditional validity with respect to distributional shift, we consider coverage conditioned on the distance score $r(x)$. Since $r(x)$ is continuous, we discretise its range into $B$ bins $\{\mathcal{B}_b\}_{b=1}^B$. Let

\begin{equation}
	\mathcal{I}_b = \{ i \in \{1,\dots,n_{\mathrm{val}}\} : r(x_i) \in \mathcal{B}_b \}
\end{equation}

denote the set of validation points whose distance scores fall into bin $\mathcal{B}_b$. We define the distance-stratified coverage as

\begin{equation}
	\xi_b = \frac{1}{|\mathcal{I}_b|} \sum_{i \in \mathcal{I}_b} \mathbb{I}\{ y_i \in C(x_i) \}.
	\label{eq:distance_coverage_metric}
\end{equation}

\textbf{Quantile-scaled predictive breadth.}
To quantify the imprecision of a p-box prediction, we measure the area between its upper and lower bounds, $\gamma(x) = \int \big( \overline{G}(t) - \underline{G}(t) \big) \, dt$, whose magnitude depends on the scale of the response variable, which varies across datasets. We further introduce a scale-adjusted version based on the empirical range of the data. 
Specifically, let $Q_{0.05}$ and $Q_{0.95}$ denote the 5\% and 95\% quantiles of the response variable. We define the quantile-scaled breadth as

\begin{equation}
	\gamma_n(x) = \frac{\gamma(x)}{Q_{0.95} - Q_{0.05}}.
\end{equation}

This scaling yields a dimensionless measure of predictive imprecision that is robust to outliers and comparable across datasets. Larger values of $\gamma_n$ indicate greater epistemic uncertainty in the prediction.

\textbf{Pooled accuracy}
To obtain an overall performance measure, we aggregate individual p-box predictions into a pooled representation using \textit{u-pooling} \cite{ferson2008model}, which generalises classical PIT-based calibration diagnostics to imprecise predictions. 
Each observation is mapped to an interval-valued $u$ via the predicted p-box. Pooling these intervals yields an empirical p-box on $[0,1]$, and calibration is quantified by the area between this p-box and the uniform distribution.
Lower values of $d_p$ indicate better calibration.

\begin{equation}
	d_p = \int_0^1 \big( \overline{G}_N(u) - \underline{G}_N(u) \big) \, du.
\end{equation}



\section{Experiments}

We present empirical evaluations of the proposed method: first on a toy dataset to illustrate extrapolation behaviour and predictive uncertainty, and then on UCI regression benchmarks to assess performance under distributional shift and different data sizes.
For CII, calibration points are sampled to span the full range of distance scores while ensuring sufficient representation near the edge of the training support, where extrapolation effects are most pronounced.


\subsection{Predictive uncertainty on the extrapolation task}

We start with a toy dataset, i.e. the cubic function regression $y = x^3 + \epsilon_n$ where $\epsilon_n \sim \mathcal{N}(0, 9)$ \cite{hernandez2015probabilistic}. 
A small sample set of 40 data points are generated from $\mathcal{U}(-4, 4)$ but the prediction is required on $[-8, 8]$.
Fig.~\ref{fig:cubic_toy_example}(a) shows the bounds from the established probabilistic predictors. 
MVE models the heteroscedastic aleatory uncertainty \cite{sluijterman2024optimal}, while MoG suggests a mixture of an ensemble of Gaussian distributions obtained through sampling the posterior weights learned from variational inference $\mathcal{E} = \{ f^{\boldsymbol{\omega}}(x), \boldsymbol{\omega} \sim p(\boldsymbol{\omega}|\mathcal{D})\}$. To maintain the epistemic uncertainty in the set and compound the aleatory and epistemic uncertainty, We take the envelope of the ensemble of distributions as the base hypothesis, the p-box $\mathcal{P}_{base} = [\underline{F}(x), \overline{F}(x)]$, into the CII framework using  Eq.~(\ref{eq:CII_framework}),
%
in which the lower and upper bounds are respectively $\underline{F}(x) = \inf_{i \in \mathcal{E}} F_i(x)$ and $\overline{F}(x) = \sup_{i \in \mathcal{E}} F_i(x)$.

As seen from Fig.~(\ref{fig:MVE_MoG_cubic}), MVE suffers from this small data regime whereas MoG produces good uncertainty estimates for interpolation but fails to rigorously enclose the observation in the extrapolation domain, indicated outside the red vertical bars.
In contrast, CII produces high-quality extrapolation, yielding an uncertainty band that not only fully encloses the observations even far beyond the training support (Fig.~\ref{fig:range_mean_band}), but also captures a consistent trend, with the upper boundary continuing to enclose the underlying function (Fig.~\ref{fig:CII_bands_cubic}).
Fig.~(\ref{fig:pbox_pred_ex_cubic}) shows the comparison between the conformalised p-box and the predictive distributions from the other two methods.


\subsection{Effects of training data size on uncertainty evaluation} 

In practice,  data sets are often expensive, unreliable or scarce. The availability of data is seen as a major source epistemic uncertainty in casting inference on data-driven models. 
To comprehensively test and demonstrate the capability of CII approach against limited data and extrapolation, we employed UCI datasets commonly used in previous endeavours \cite{gal2016dropout} of quantifying predictive uncertainty. 
However, it should be noted that, for the particular objective of evaluating extrapolation performance, we partition the dataset into training and holdout subsets via quantile thresholds of the distance score in a ratio of 3:7. This splitting process is repeated 20 times and the average test performance of  each method is reported.


In addition, to investigate the impact of dataset size on the learned uncertainty and the resulting predictions, we conduct experiments across varying proportions of the training data, i.e. $\epsilon=\{0.1, 0.3, 0.7, 0.9\}$, of the training data.

\begin{figure*}[t]
	\centering
	\includegraphics[width=\textwidth]{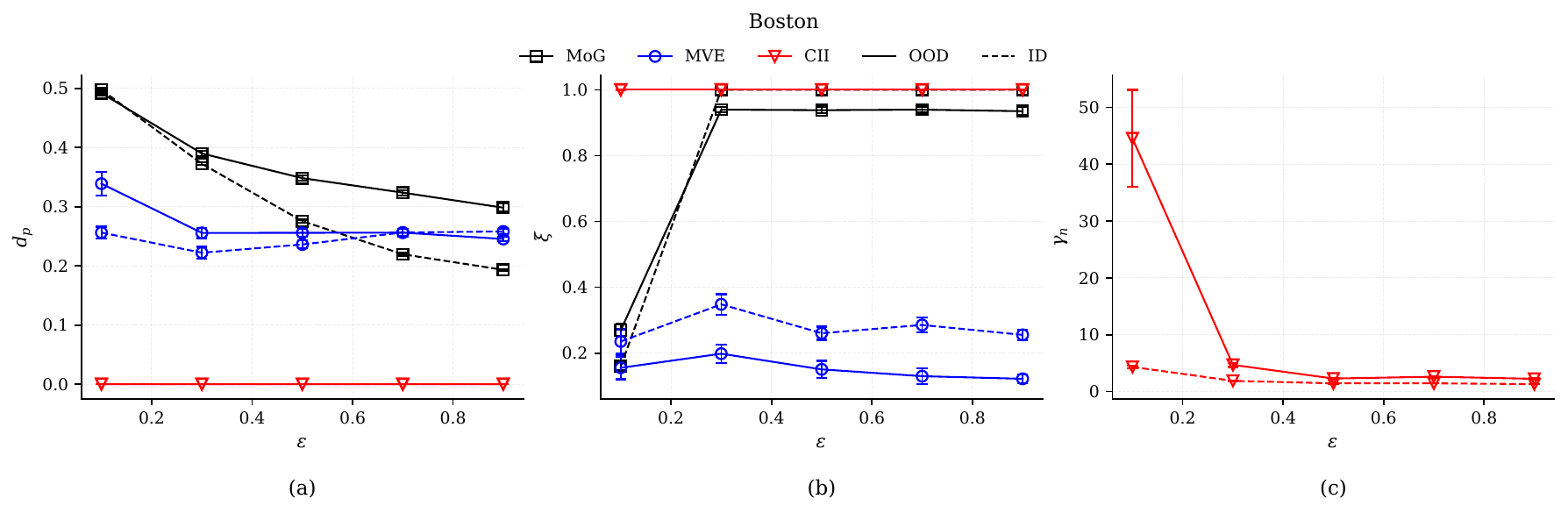}
	\caption{Performance comparison on the Boston dataset under varying data proportions $\epsilon$. Panels (a)–(c) report pooled calibration error $d_p$, coverage $\xi$, and quantile-scaled breadth $\gamma_q$, respectively. Results are shown for in-distribution (ID, dashed) and out-of-distribution (OOD, solid) settings.}
	\label{fig:dataset_3panel_summary_boston}
\end{figure*}


Fig.~\ref{fig:dataset_3panel_summary_boston} summarises performance on the Boston dataset across varying data proportions. The proposed CII method consistently achieves the lowest pooled discrepancy error $d_p$ while maintaining perfect empirical coverage $\xi$ in both in-distribution (ID) and out-of-distribution (OOD) settings. This indicates reliable and well-calibrated predictions across data regimes.

In contrast, MoG attains high coverage in the ID setting but exhibits degradation under distribution shift, particularly at small data proportions, as reflected by increased $d_p$ and reduced OOD coverage. MVE performs substantially worse, with consistently low coverage due to the absence of epistemic uncertainty modelling.

These observations are consistent with the cross-dataset results in Table~\ref{tab:cross_dataset_summary}, where CII uniformly preserves coverage while adapting predictive imprecision. In particular, the increase in $\gamma_n$ under OOD settings (Fig.~\ref{fig:dataset_3panel_summary_boston}(c)) aligns with the broader trend that CII expands uncertainty under distributional shift, whereas baseline methods fail to maintain validity. Due to limit space, only the metrics associated with $\epsilon=0.5$ is reported in Table~\ref{tab:cross_dataset_summary}.

\begin{table*}[t]
\centering
\footnotesize
\setlength{\tabcolsep}{3pt}
\renewcommand{\arraystretch}{0.92}
\caption{Cross-dataset performance at fixed $\epsilon=0.5$ for in-distribution (ID) and out-of-distribution (OOD) evaluation.}
\resizebox{\textwidth}{!}{%
\begin{tabular}{llclllllll}
\toprule
Dataset & $\epsilon$ & Distribution
& \multicolumn{2}{c}{MoG}
& \multicolumn{2}{c}{MVE}
& \multicolumn{3}{c}{CII} \\
\cmidrule(lr){4-5}
\cmidrule(lr){6-7}
\cmidrule(lr){8-10}
& & 
& $d_p$ & $\xi$
& $d_p$ & $\xi$
& $d_p$ & $\xi$ & $\gamma_n$ \\
\midrule
Boston & 0.5 & ID & $0.28 \pm 0.00$ & $\mathbf{1.00 \pm 0.00}$ & $0.24 \pm 0.01$ & $0.26 \pm 0.02$ & $\mathbf{0.00 \pm 0.00}$ & $\mathbf{1.00 \pm 0.00}$ & $1.45 \pm 0.03$ \\
 &  & OOD & $0.35 \pm 0.00$ & $0.94 \pm 0.01$ & $0.26 \pm 0.01$ & $0.15 \pm 0.03$ & $\mathbf{0.00 \pm 0.00}$ & $\mathbf{1.00 \pm 0.00}$ & $2.30 \pm 0.14$ \\
Concrete & 0.5 & ID & $0.16 \pm 0.01$ & $\mathbf{1.00 \pm 0.00}$ & $0.13 \pm 0.01$ & $0.70 \pm 0.02$ & $\mathbf{0.00 \pm 0.00}$ & $\mathbf{1.00 \pm 0.00}$ & $2.49 \pm 0.06$ \\
 &  & OOD & $0.25 \pm 0.01$ & $0.99 \pm 0.00$ & $0.17 \pm 0.01$ & $0.62 \pm 0.03$ & $\mathbf{0.00 \pm 0.00}$ & $\mathbf{1.00 \pm 0.00}$ & $13.02 \pm 1.58$ \\
Energy & 0.5 & ID & $0.20 \pm 0.00$ & $\mathbf{1.00 \pm 0.00}$ & $0.14 \pm 0.01$ & $0.77 \pm 0.02$ & $\mathbf{0.00 \pm 0.00}$ & $\mathbf{1.00 \pm 0.00}$ & $3.32 \pm 0.07$ \\
 &  & OOD & $0.19 \pm 0.00$ & $\mathbf{1.00 \pm 0.00}$ & $0.16 \pm 0.01$ & $0.62 \pm 0.03$ & $\mathbf{0.00 \pm 0.00}$ & $\mathbf{1.00 \pm 0.00}$ & $6.41 \pm 0.41$ \\
Kin8nm & 0.5 & ID & $0.16 \pm 0.00$ & $\mathbf{1.00 \pm 0.00}$ & $0.11 \pm 0.01$ & $0.97 \pm 0.01$ & $\mathbf{0.00 \pm 0.00}$ & $\mathbf{1.00 \pm 0.00}$ & $4.05 \pm 0.05$ \\
 &  & OOD & $0.15 \pm 0.00$ & $\mathbf{1.00 \pm 0.00}$ & $0.12 \pm 0.01$ & $0.86 \pm 0.02$ & $\mathbf{0.00 \pm 0.00}$ & $\mathbf{1.00 \pm 0.00}$ & $4.39 \pm 0.07$ \\
Naval Propulsion & 0.5 & ID & $0.24 \pm 0.00$ & $\mathbf{1.00 \pm 0.00}$ & $0.28 \pm 0.02$ & $\mathbf{1.00 \pm 0.00}$ & $\mathbf{0.00 \pm 0.00}$ & $\mathbf{1.00 \pm 0.00}$ & $234.64 \pm 2.86$ \\
 &  & OOD & $0.24 \pm 0.00$ & $\mathbf{1.00 \pm 0.00}$ & $0.35 \pm 0.03$ & $0.50 \pm 0.06$ & $\mathbf{0.00 \pm 0.00}$ & $\mathbf{1.00 \pm 0.00}$ & $241.62 \pm 3.22$ \\
Power Plant & 0.5 & ID & $0.45 \pm 0.00$ & $\mathbf{1.00 \pm 0.00}$ & $0.06 \pm 0.00$ & $\mathbf{1.00 \pm 0.00}$ & $\mathbf{0.00 \pm 0.00}$ & $\mathbf{1.00 \pm 0.00}$ & $65.36 \pm 0.06$ \\
 &  & OOD & $0.46 \pm 0.00$ & $\mathbf{1.00 \pm 0.00}$ & $0.30 \pm 0.01$ & $0.73 \pm 0.02$ & $\mathbf{0.00 \pm 0.00}$ & $\mathbf{1.00 \pm 0.00}$ & $74.95 \pm 0.49$ \\
Wine & 0.5 & ID & $0.13 \pm 0.00$ & $\mathbf{1.00 \pm 0.00}$ & $0.16 \pm 0.01$ & $0.54 \pm 0.02$ & $\mathbf{0.00 \pm 0.00}$ & $\mathbf{1.00 \pm 0.00}$ & $3.39 \pm 0.09$ \\
 &  & OOD & $0.11 \pm 0.00$ & $\mathbf{1.00 \pm 0.00}$ & $0.19 \pm 0.01$ & $0.45 \pm 0.02$ & $\mathbf{0.00 \pm 0.00}$ & $\mathbf{1.00 \pm 0.00}$ & $8.63 \pm 0.86$ \\
Yacht & 0.5 & ID & $0.13 \pm 0.00$ & $0.89 \pm 0.01$ & $0.18 \pm 0.01$ & $0.69 \pm 0.03$ & $\mathbf{0.01 \pm 0.00}$ & $\mathbf{1.00 \pm 0.00}$ & $0.69 \pm 0.04$ \\
 &  & OOD & $0.12 \pm 0.00$ & $0.89 \pm 0.01$ & $0.18 \pm 0.01$ & $0.46 \pm 0.02$ & $\mathbf{0.00 \pm 0.00}$ & $\mathbf{1.00 \pm 0.00}$ & $1.19 \pm 0.11$ \\
\bottomrule
\end{tabular}
}
\label{tab:cross_dataset_summary}
\end{table*}

\begin{figure}[h!]
	\centering
	\includegraphics[width=0.5\textwidth]{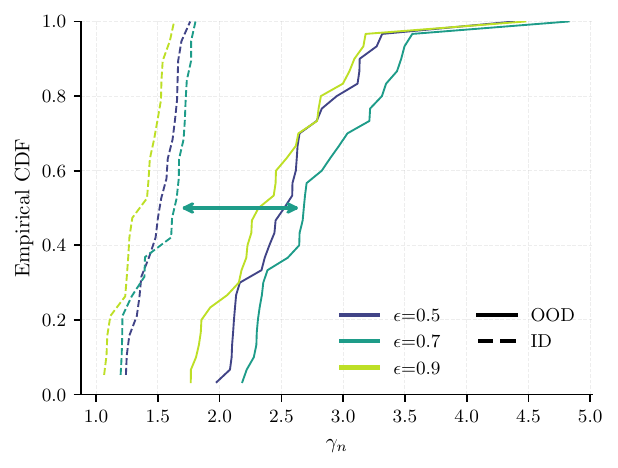}
	\caption{Empirical CDFs of the quantile-scaled predictive breadth $\gamma_n$ for different training proportions $\epsilon$ in the Boston dataset. Solid lines correspond to OOD data, while dashed lines correspond to ID data.}
	\label{fig:ID_vs_OOD}
\end{figure}

Fig.~\ref{fig:dataset_3panel_summary_boston}(c) illustrates the increase in predictive imprecision of the conformalised p-box as the available data decreases. With fewer training samples, model uncertainty grows, leading to broader predictive sets. Notably, a clear separation emerges between the ID and OOD regimes, with OOD predictions exhibiting consistently larger imprecision. This distinction is further highlighted in Fig.~(\ref{fig:ID_vs_OOD}).

This awareness of distribution shift is illustrated in Fig.~(\ref{fig:ID_vs_OOD}), where the empirical CDFs of the quantile-scaled breadth $\gamma_n$ exhibit a systematic gap (a rightward shift) from ID (dashed) to the OOD (solid) regime. 
For a fixed data proportion $\epsilon$, the OOD curves consistently indicate larger predictive breadth, reflecting increased uncertainty under distribution shift. Importantly, this adjustment occurs while maintaining valid coverage, demonstrating that the proposed method encodes distributional mismatch directly into the geometry of the predictive sets.

Importantly, this distance is not used as a hard classifier alone, but serves as a continuous variable that drives a second-order model $\mathcal{M}(r)$, which captures the growth of model discrepancy and predictive uncertainty as inputs move away from the training support. By working with this geometry-aware scalar, we obtain a simple yet effective mechanism to characterise extrapolation and modulate uncertainty in a principled manner. 

\begin{figure}[h!]
	\centering
	\includegraphics[width=0.5\columnwidth]{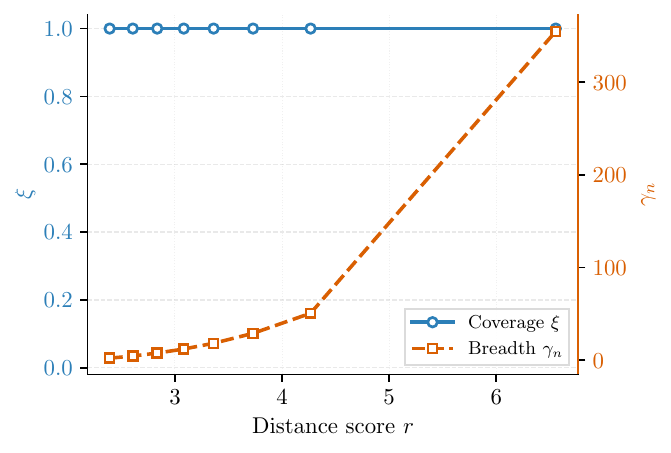}
\caption{
Distance-stratified coverage $\xi_b$ and quantile-scaled breadth $\gamma_q$ versus the distance score $r(x)$ in the wine dataset. In this experiment, no proportion is used to maintain the number of samples at distances.}
	\label{fig:coverage_breadth_vs_distance}
\end{figure}

Fig.~\ref{fig:coverage_breadth_vs_distance} investigates the \textit{adaptivity} of the proposed conformalised approach based on the distance-stratified coverage, see Eq.~(\ref{eq:distance_coverage_metric}). That is, it is desired to return larger prediction sets for difficult inputs. 

While $\xi_b$ remains approximately constant across distance bins, indicating 
robust validity, the predictive breadth $\gamma_n$ increases with $r(x)$, capturing the growth of epistemic uncertainty under extrapolation. This highlights the effectiveness of distance-aware calibration 
in maintaining coverage while adapting uncertainty to distributional shift.

\begin{figure}[h!]
	\centering
	\includegraphics[width=0.5\columnwidth]{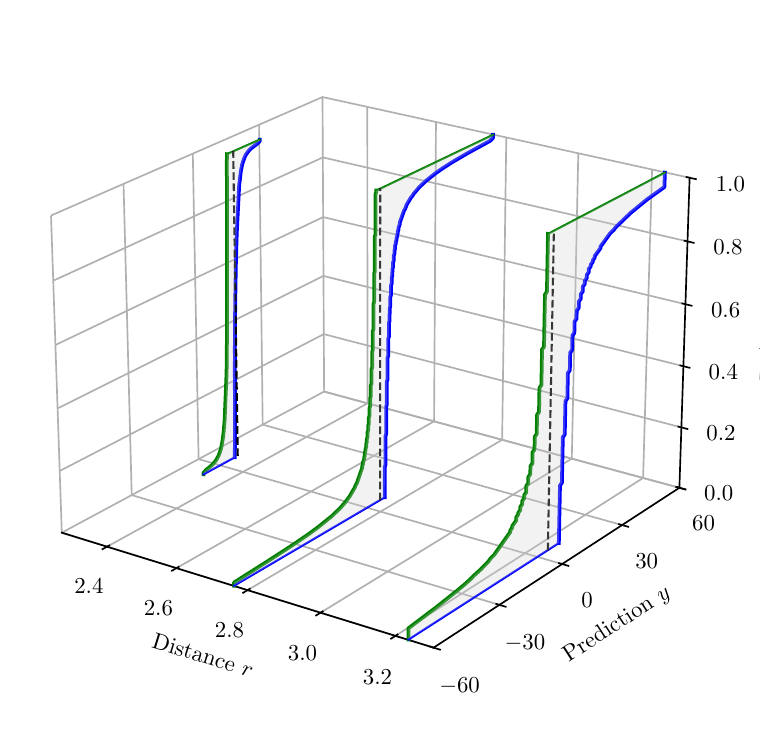}
	\caption{Difference of imprecision in conformalised predictions along with distance. Observation is indicated as the black dotted line.}
	\label{fig:pbox_w_r}
\end{figure}

Further to Fig.~(\ref{fig:coverage_breadth_vs_distance}), Fig.~(\ref{fig:pbox_w_r}) displays more intuitively the breadth of the comformalised prediction, which clearly shows the inflation of predictive uncertainty along with the distance.
Besides, CII has perfect coverage which delivers rigorous predictions that fully enclose the observation.

\section{Conclusion and outlook}

We proposed a conformalised imprecise inference framework for robust uncertainty quantification under extrapolation. 
A distance-aware calibration mechanism is used to characterise deviation from the training data manifold and to drive an uncertainty-aware discrepancy model.
The method produces prediction sets (probability boxes), realised through the \textit{ascloseas} bounds,  that maintain coverage while adaptively expanding uncertainty under distributional shift. Empirical results demonstrate consistent robustness across datasets and data regimes, particularly in small-data settings. These findings highlight the importance of incorporating distributional awareness into predictive uncertainty.

The mechanism of probability bounding provides a natural way to integrate various sources of epistemic uncertainty through inflating the bounds. The existing framework already has some capacity in dealing measurement uncertainty but not rigorously investigated. Future work will explore extensions that more comprehensively address additional epistemic uncertainties including the measurement uncertainty and alternative discrepancy models for improved scalability.

\newpage
\appendix
\onecolumn
\section{Stochastic area metric and u-pooling}

Stochastic area metric stands for a unified discrepancy measure between two uncertain numbers, which could be embodied as several uncertainty constructs including intervals, probability distributions, probability boxes (p-boxes) and Dempster-Shafer structures (DSS), plus real numbers. Both the model predictions and the observation could be subject to both aleatory and epistemic uncertainty.

For each observation $y_i$, the predicted p-box $[F_i^-, F_i^+]$ induces an interval-valued $u$-score:
\begin{equation}
	U_i = [F_i^-(y_i),\, F_i^+(y_i)] \subset [0,1].
\end{equation}

This defines a set-valued mapping from observations to the unit interval. Collecting $\{U_i\}_{i=1}^N$ yields an empirical random set on $[0,1]$, which can be represented by its lower and upper empirical distribution functions
\begin{equation}
\underline{G}_N(u) = \frac{1}{N}\sum_{i=1}^N \mathbf{1}\{F_i^+(y_i) \le u\}, 
\qquad
\overline{G}_N(u) = \frac{1}{N}\sum_{i=1}^N \mathbf{1}\{F_i^-(y_i) \le u\}.	
\end{equation}

These define a p-box $[\underline{G}_N, \overline{G}_N]$ on $[0,1]$, which characterises the pooled uncertainty in probability space. Under perfect calibration, this p-box collapses to the uniform distribution.

We quantify deviation from calibration by measuring the imprecision of the pooled $u$-distribution:

\begin{equation}
	d_p = \int_0^1 \big( \overline{G}_N(u) - \underline{G}_N(u) \big)\,du,
\end{equation}

which corresponds to the area between the lower and upper bounds of the aggregated p-box. Smaller values of $d_p$ indicate better calibration, with $d_p = 0$ attained when all $u$-scores are degenerate and uniformly distributed.


\bibliographystyle{unsrt}  
\bibliography{references}

\end{document}